\documentclass{article}

 \PassOptionsToPackage{numbers, compress}{natbib}

 \usepackage[final]{nips_2018}

\newcommand*\samethanks[1][\value{footnote}]{\footnotemark[#1]}

\usepackage{subcaption} 
\usepackage[utf8]{inputenc} 
\usepackage[T1]{fontenc}    
\usepackage{hyperref}       
\usepackage{url}            
\usepackage{booktabs}       
\usepackage{amsfonts}       
\usepackage{nicefrac}
\usepackage{amsmath}
\usepackage{graphicx}       
\usepackage{microtype}   
\usepackage{algorithmic, algorithm}

\DeclareMathOperator{\E}{E}

\graphicspath{ {figures/} }
\title{Black-Box Autoregressive Density Estimation for State-Space Models}

\author{
	Tom Ryder\thanks{Equal contribution}, Andrew Golighty, A. Stephen McGough, Dennis Prangle\samethanks\\
	Newcastle Univeristy, Newcastle, UK\\
	\texttt{\{t.ryder2, dennis.prangle\}@newcastle.ac.uk} 
}

\begin{document}

\maketitle
%

\section{Introduction}
State-space models (SSMs) provide a flexible framework for modelling time-series data.
Consequently, SSMs are ubiquitously applied in areas such as engineering \cite{elliott2008hidden}, econometrics \cite{Black:1973} and epidemiology \cite{Fuchs:2013}.
In this paper we provide a fast approach for approximate Bayesian inference in SSMs
 using the tools of deep learning and variational inference.

Formally, a SSM is based on a latent Markov process $X_{t_i}$ at times $t_i = 0, \Delta t, 2\Delta t, \ldots, N$ for some $\Delta t > 0$.
The SSM has initial density $p(x_{t_0})$ and evolves through a \textit{transition density}
$
X_{t_i} | \left(X_{t_{i-1}} = x_{t_{i-1}}\right) \sim p(x_{t_i} | x_{t_{i-1}}, \theta).
$
Observations $Y_{t_i}$ of the latent process are available according to an \textit{observation likelihood}
$
Y_{t_i} | \left(X_{t_i} = x_t\right) \sim p(y_{t_i} | x_{t_i}, \theta).
$
Here ${\theta}$ denotes the set of \textit{global} latent variables that govern the above densities.
\paragraph{Bayesian Inference} We will operate in a Bayesian framework where, after ascribing prior densities $p(\theta)$ and $p(x_{t_0})$, interest lies in the posterior density 
\begin{equation}\label{eq:bayes_SS}
p\left(x_{t_0:t_N},\theta | y_{t_0:t_N}\right) \propto p(\theta) p(x_{t_0})  \prod_{i = 1}^{N}   p(x_{t_i} | x_{t_{i-1}}, \theta) \prod_{i = 0}^{N} p(y_{t_i} | x_{t_i}, \theta).
\end{equation}
A popular approach to Bayesian inference is the use of sampling techniques such as particle filtering and Markov chain Monte Carlo \cite{Doucet2001, Capp:2010:IHM:1965046}. These methods, however, do not typically scale well to large datasets and can be inefficient when only partial and/or sparse observations of the latent process are available (see Appendix \ref{sec:PartialObs} for details). A promising alternative to sampling is variational inference. Here we introduce a family of approximations to the posterior and select the member closest to the true posterior. The approximate family is often chosen such that its parameters are differentiable with respect to the objective, permitting optimization with  stochastic gradient descent. This technique subsumes a broad class of methods known as \textit{black-box }variational inference (BBVI)\cite{pmlr-v33-ranganath14}.

\paragraph{Related Work and Contribution}
Several recent authors have looked at BBVI for SSMs. This work can be broadly separated into: a. approaches proposing forms of variational approximation for SSMs (e.g. \cite{pmlr-v80-ryder18a, Archer:2016, pmlr-v80-binkowski18a}); b. approaches developing tighter bounds on the evidence e.g.~using ideas from sequential Monte Carlo \cite{pmlr-v84-naesseth18a,le2017auto,DBLP:conf/nips/MaddisonLTHNMDT17}. Our contribution is to introduce a variational approximation based on modern autoregressive density estimators. This approach, which exploits the speed of GPU computation, is extremely fast and flexible enough to produce a close approximation to the joint posterior for $(\theta, x)$.
 

\section{Approximate Bayesian Inference}

\paragraph{Variational Inference} Variational inference (see e.g.~\citealp{Blei:2017}) recasts the numerical integration problem of posterior inference \eqref{eq:bayes_SS} as one of \textit{optimization}. Inference then proceeds by introducing a family of approximations to the posterior, $q(x_{t_0:t_N}, {\theta}; {\phi})$, and minimising the Kullback-Leibler divergence $KL[q(x_{t_0:t_N}, {\theta}; {\phi}) || p({x_{t_0:t_N}}, {\theta} | {y_{t_0:t_N}})]$ with respect to the collection of variational parameters ${\phi} $. This is equivalent to maximising the ELBO (evidence lower bound) \cite{Jordan1999},
\begin{equation} \label{eq:ELBO}
\E_{{x}, {\theta} \sim q(\cdot; \phi)} [ \log p({x_{t_0:t_N}}, {y_{t_0:t_N}}, {\theta}) - \log q({x_{t_0:t_N}}, {\theta}; {\phi})].
\end{equation}
The optimal $q({x_{t_0:t_N}}, {\theta}; {\phi})$ is an approximation to the posterior distribution.
This is typically overconcentrated, unless the approximating family allows particularly close matches to the posterior.

\paragraph{Inverse Autoregressive Flows}
The approximation error of variational inference can be alleviated by using a highly flexible approximate posterior. A key research theme has been designing expressive densities that remain computationally tractable (e.g. \cite{journals/corr/KingmaW13, Bengio00modelinghigh-dimensional, pmlr-v15-larochelle11a,pmlr-v37-germain15, NIPS2013_5060,Papamakarios:2017:maf}). Of particular interest here is work on \textit{normalising flows} \cite{Rezende:2015:VIN:3045118.3045281} and \textit{inverse autoregressive flows} (IAFs) \cite{DBLP:journals/corr/KingmaSW16}.

A normalising flow represents a random variable $x$ as $g(z)$: a learnable bijection of a base random variable $z$. Typically $z \sim N(0,I)$.
An IAF specifies
\begin{equation} \label{eq:IAF}
x_i = \mu_i(z_{1:{i-1}}) + \sigma_i(z_{1:{i-1}}) z_i.
\end{equation}
An IAF is flexible and, when $\dim(x)$ is small, allows fast sampling -- e.g.~using GPUs -- and fast calculation of a sample's log density.
Also, using IAFs for $q$ in \eqref{eq:ELBO} allows gradient estimates to be calculated using automatic differentiation \cite{journals/corr/KingmaW13, Rezende:2015:VIN:3045118.3045281, icml2014c2_titsias14}.
Hence IAFs are well suited for variational inference.
It is common for the $\mu$ and $\sigma$ functions to be neural network outputs with learnable parameters $\phi$.
See \cite{Papamakarios:2017:maf} for an efficent scheme requiring only a single neural network.

Typically several IAF transformations, optionally separated by permutation operations, are composed to give the overall variational density.


\paragraph{Black-Box Autoregressive Density Estimation for SSMs}
IAFs become expensive for high $\dim(x)$ due to the large number of inputs to the $\mu_i$ and $\sigma_i$ functions.
We introduce a \emph{local IAF} of a similar form to Wavenet \cite{journals/corr/OordDZSVGKSK16},
\begin{equation} \label{eq:localIAF}
x_i = \mu(z_{i-k:i-1}) + \sigma(z_{i-k:i-1}) z_i,
\end{equation}
(where we use padding to deal with $z_i$ values with $i<0$.)
Here the mean and variance depend only on a \emph{local receptive field} of length $k$.
This is suitable for SSMs whose posteriors exhibit short-range dependence.
Note that the $\mu$ and $\sigma$ sequences can be seen as outputs of a 1D convolutional neural network with an off-centre receptive field.
This amortizes the cost of inference for $x$.

Our variational approximation to the posterior \eqref{eq:bayes_SS} is
\begin{equation} \label{eq:q}
q(\theta, x; \phi) = q(\theta; \phi_{\theta}) q(x | \theta ; \phi_{x}),
\end{equation}
where $\phi_x$ and $\phi_{\theta}$ represent the weights of the neural networks used to approximate $x$ and $\theta$, respectively.
For $q(\theta; \phi_{\theta})$ we use several composed IAFs based on \cite{Papamakarios:2017:maf} with random permutations.
Our $ q(x | \theta ; \phi_{x})$ uses composed local IAFs and order-reversing permutations, and, where necessary, a final transformation constraining $x$ to positive values.
These local IAFs also include a dependence of $\mu$ and $\sigma$ on $\theta$ and data features from $y_{i-k:i-1}$.

We optimize the ELBO using standard stochastic gradient methods. Additionally, we use tempering to encourage better exploration of the ${\theta}$ space, replacing $q(\theta; \phi_{\theta})$ with $q(\theta; \phi_{\theta})^\alpha$ in the ELBO and reducing $\alpha$ from a large initial value to 1 during training.

See Appendix \ref{sec:fullMethod} for further details of our variational approximation and optimization.

\section{Experiments}
\paragraph{Diffusion Processes}
As a special case of a latent-variable state-space model, consider the $p$-dimensional It\^o process $\{X_{t}\}_{_{t\geq 0}}$ satisfying the stochastic differential equation (SDE) 
\begin{equation} \label{eq:SDEs}
dX_t = \alpha(X_t, {\theta}) dt + \sqrt{\beta(X_t, {\theta})} dW_t, \quad X_0 = x_0,
\end{equation}
together with the simple additive Gaussian observation model
\begin{equation}\label{eq:obs_model}
Y_{t_i} = F'X_{t_i}+ \epsilon_{t_i}, \quad \epsilon_{t_i} \overset{indep}{\sim } N(0, \sigma^2 I).
\end{equation}
Here $\alpha$ is a $p$-dimensional \emph{drift vector},  $\beta$ is a $p \times p$ positive definite \emph{diffusion matrix} (with $\sqrt{\beta}$ representing a matrix square root), $W_t$ is a $p$-vector of standard and uncorrelated Brownian motion processes, $F$ is a constant $p \times p_0$ matrix and $\sigma^2$ is the variance of the observation error, which may be assumed known or the object of inference. For the latter case $\sigma$ should be a specified function of ${\theta}$.

\paragraph{Discretisation}
Few SDEs permit analytical solutions and it is common to rely on an approximate transition density based on a time discretisation.
For our purpose, we work with the Euler-Maruyama scheme, in which transitions between states at successive times are approximated as Gaussian so that
\begin{equation} \label{eq:EMT}
p \left(x_{t_i} | x_{t_{i-1}}, {\theta}\right) = N \big(x_{t_{i}}-x_{t_{i-1}}; \, \alpha(x_{t_{i-1}}, {\theta}) \Delta t, \beta(x_{t_{i-1}}, {\theta}) \Delta t\big),
\end{equation}
where, as defined earlier, $\Delta t = t_i - t_{i-1}$, the time between successive latent values.
\paragraph{Ornstein-Uhlenbeck}
As a simple illustration, we begin by implementing our method for the univariate, mean-reverting Ornstein-Uhlenbeck process governed by the following SDE
\begin{equation}\label{OU_SDE}
dX_t = \theta_1 \left(\theta_2 - X_{t}\right) dt + \theta_3 dW_t,
\end{equation}
where ${\theta} = (\theta_1, \theta_2, \theta_3)'$. Unlike most SDEs the Ornstein-Uhlenbeck process \eqref{eq:obs_model} permits a closed-form solution. It is therefore possible to recover the exact posterior for our global parameters ${\theta}$ for direct comparison with our variational approach using a simple forward filter recursion (see Appendix \ref{sec:forward_filter}). 

\begin{figure}
 	\centering
 	\includegraphics[width=0.88\textwidth]{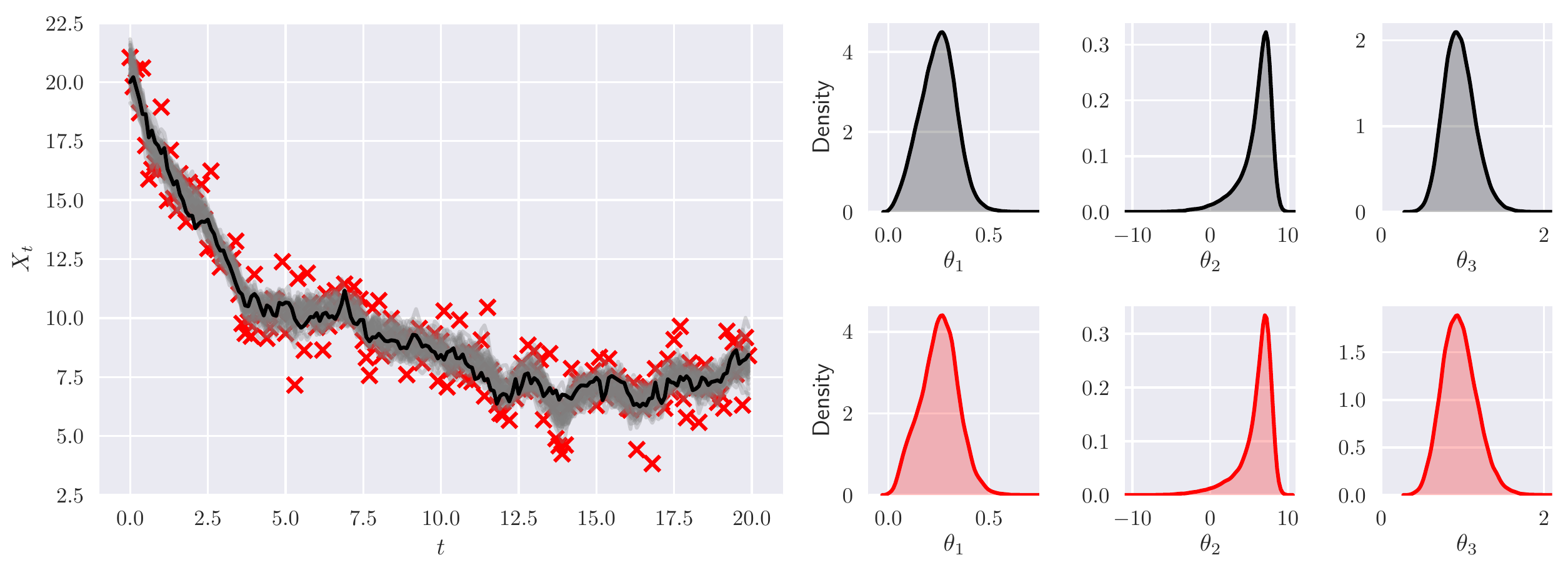}
	\includegraphics[width=0.88\textwidth]{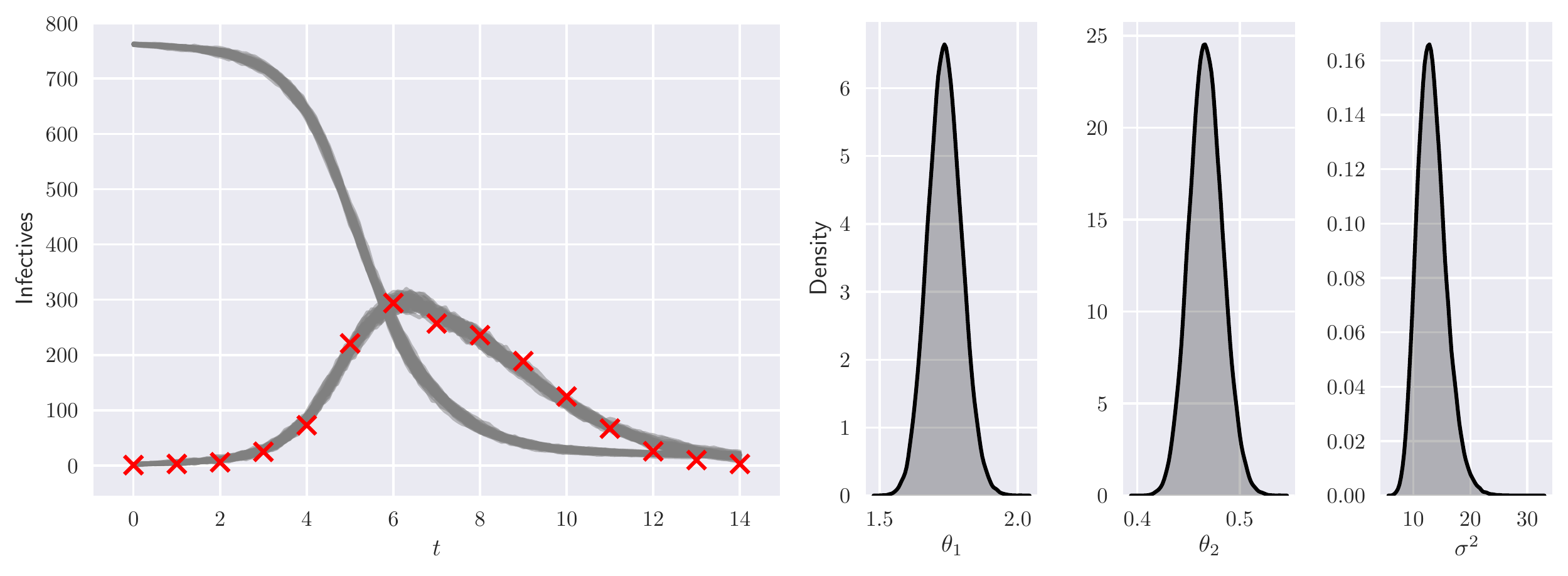}
 	\caption{Top: Ornstein-Uhlenbeck example. Bottom: epidemic example. Left: 50 samples from the approximate smoothing density (grey), and noisy observations (red crosses) of the latent process (black, when available). Right: approximate marginal parameter (black) and exact posteriors from forward filter recursion (when available, red).}
 	\label{fig:1}
 \end{figure}

By using the exact solution of \eqref{OU_SDE}, with $\Delta t = 0.1$, $\theta = (0.2, 5.0, 1.0)'$, $x_0 = 20$ and $\sigma^2 =1$ (assumed known), we simulate 200 synthetic observations on the interval $[0,20]$. We then infer the partially log-transformed parameters ${\vartheta} = (\log \theta_1, \theta_2, \log\theta_3)'$ under independent $N(0, 10^2)$ priors. We implement our approach on an NVIDIA Titan XP, for which convergence took $\sim 5$  minutes. Figure \ref{fig:1} displays the variational posterior, illustrating a very close match to the exact $\theta$ marginals.

\paragraph{Epidemic Model}
An SIR epidemic model \cite{Andersson:2000} describes the spread of an infectious disease.
Here the population is subdivided into those susceptible ($S$), those infectious ($I$) and removed individuals ($R$).
For our example, we assume a hermetic population and as such only model $S_t$ and $I_t$.

Our data is on an outbreak of influenza at a boys boarding school in 1978 \cite{Jacksone002149}.
Of 763 boys at the school, 512 were infected within 14 days.
Observations of the number infectious are provided daily by those students confined to bed.
We replicate the SDE model and priors of \cite{pmlr-v80-ryder18a} (including use of a fixed $x_0$).
Figure \ref{fig:1} shows the variational posterior.
Our results are almost identical to the variational approach of \cite{pmlr-v80-ryder18a}, but are obtained much faster: convergence took only $\sim 20$ minutes rather than hours. 

Further implementation details for both examples are available in Appendix \ref{sec:imp_details}.

\subsubsection*{Acknowledgments}

Tom Ryder is supported by the Engineering and Physical Sciences Research Council, Centre for Doctoral Training in Cloud Computing for Big Data (grant number EP/L015358/1).

We acknowledge with thanks an NVIDIA academic GPU grant for this project.

\bibliographystyle{unsrt}
\bibliography{NIPS_workshop}

\appendix
\section{Case of Sparse Observations} \label{sec:PartialObs}
For completeness, consider the case of time-sparse observations and the corresponding set of observation times $S \subseteq \{t_0,t_1,t_2, \ldots, t_N\}$. In such a case, \eqref{eq:bayes_SS} becomes
\begin{equation}\label{eq:bayes_SS_partial}
p\left(x_{t_0:t_N},\theta | y_{t_0:t_N}\right) \propto p(\theta) p(x_{t_0})  \prod_{i = 1}^{N}   p(x_{t_i} | x_{t_{i-1}}, \theta) \prod_{j \in S}^{} p(y_{j} | x_{j}, \theta),
\end{equation}
and later derivations in the paper change similarly.

\section{Variational Approximation and Optimization} \label{sec:fullMethod}

 We use a composition of $m$ local IAFs, separated by order-reversing permutations, to build our $q(x | \theta; \phi_x)$ density.
This appendix describes the details and the optimization objective that results.
We'll concentrate on the case where $x_{t_0}$ is fixed and we need a variational density for $x_{t_1:t_N}$.
Both the examples in the paper are of this form.

We begin by introducing IID $N(0, 1)$ variables $z^{0}_{t_1}, z^{0}_{t_2}, \ldots , z^{0}_{t_N}$ and define, following \cite{DBLP:journals/corr/KingmaSW16} for numerical stability,
 \begin{equation}
 z_{t_i}^{j+1} = z_{t_i}^{j} \sigma_{t_i}^{j+1} + \mu_{t_i}^{j+1} (1 -  \sigma_{t_i}^{j+1})
 \end{equation}
 where if $j$ is odd
 \begin{align}
 \mu_{t_i}^{j} &= \mu^{j}\left(z_{t_{i-k}:t_{i-1}}^{j-1}, y_{t_{i-k}:t_{i-1}}, {\theta}; {\phi}_{x}^j\right), \\ \sigma_{t_i}^{j} &= \sigma^{j}\left(z_{t_{i-k}:t_{i-1}}^{j-1}, y_{t_{ i-k}:t_{i-1}}, {\theta}; {\phi}_{x}^j\right),
 \end{align}
and we replace the indices $t_{i-k}:t_{i-1}$ with $t_{i-1}:t_{i-k}$ if $j$ is even.
(This is a notationally simple way to introduce order-reversing permutations.)
We implement the functions $\mu^{j}$ and $\sigma^{j}$ through a neural network,
using the sigmoid function to ensure the output for $\sigma^{j}$ is in the interval $[0,1]$.

The equations above sometimes require $z_{t_i}^{j-1}$ and $y_{t_i}$ inputs with $i < 0$ or $i > N$
i.e.~outside the grid of times for the SSM.
To allow such inputs we assume they are all zero, effectively \emph{padding} our inputs as is often done for convolutional neural networks.

The transformation outlined above outputs $z^m_{t_1}, z^m_{t_2}, \ldots, z^m_{t_N}$.
As explained in \cite{DBLP:journals/corr/KingmaSW16}, the corresponding Jacobian is the product of all the $\sigma_{t_i}^j$ terms.
We apply a final elementwise transformation $h$ to give output $x_{t_1}, x_{t_2}, \ldots, x_{t_N}$.
In our examples we take $h$ to be the softplus function when $x$ is required to be positive and otherwise we use the identity.
The overall density is
\begin{equation} \label{eq:qx}
q(x | \theta; \phi_x) = \frac{\prod_{i=1}^N p(z_{t_i}^{(0)})}{\prod_{i=1}^N h'(z^m_{t_i}) \prod_{i=1}^N \prod_{j=1}^m \sigma^j_{t_i}},
\end{equation}
where $p(z_{t_i}^{(0)})$ is a $N(0,1)$ density.

Using \eqref{eq:qx} in our variational approximation \eqref{eq:q} gives the ELBO
 \begin{equation} \label{eq:ELBO3}
 \begin{split}
 \mathcal{L}(\phi) = \E_{{{\theta}}, {x} \sim  q(\cdot; \phi)}&\Bigg[\log p({\theta}) - \log q({\theta}; \phi_\theta) +\sum_{i=1}^{N} \Big\{   \log p(x_{t_i} | x_{t_{i-1}}, {\theta})\\ & + \log p(y_{t_i} | x_{t_i}, {\theta})  - p(z_{t_i}^0) + h'(z_{t_i}^{m}) + \sum_{j = 1}^{m} \log \sigma_{t_i}^j \Big\} \Bigg].
 \end{split}
 \end{equation}
We now apply the reparameterisation trick \cite{icml2014c2_titsias14, Rezende:2015:VIN:3045118.3045281, journals/corr/KingmaW13}.
We have defined $q$ so that $\theta,x$ is a transformation of a vector $z^0$ of IID $N(0,1)$ random variables.
Hence \eqref{eq:ELBO3} can be represented as an expectation over $z^0$, and easily differentiated with respect to $\phi$.
So an unbiased Monte Carlo estimate of $\nabla_\phi \mathcal{L}(\phi)$ is
  \begin{equation} \label{eq:ELBO4}
 \begin{split}
\widehat{\nabla_\phi \mathcal{L}(\phi)} = \frac{1}{n}\sum_{\ell=1}^{n}&\nabla_\phi\Bigg[\log p({\theta^{(\ell)}}) - \log q({\theta^{(\ell)}}; \phi_\theta) +\sum_{i=1}^{N} \Big\{   \log p(x_{t_i}^{(\ell)} | x_{t_{i-1}}^{(\ell)}, {\theta}^{(\ell)})\\ & + \log p(y_{t_i} | x_{t_i}^{(\ell)}, {\theta}^{(\ell)})  - p(z_{t_i}^{0, (\ell)}) + h'(z_{t_i}^{m, (\ell)}) + \sum_{j = 1}^{m} \log \sigma_{t_i}^{j, (\ell)} \Big\} \Bigg],
 \end{split}
 \end{equation}
where each $(\theta^{(\ell)}, x^{(\ell)})$ is based on an independent $z^0$ sample.
The right hand side of \eqref{eq:ELBO4} can be calculated using automatic differentiation, and the resulting $\nabla_\phi \mathcal{L}(\phi)$ estimates used in stochastic gradient descent.

As mentioned in the main text, we also used a tempering approach, replacing $q(\theta^{(\ell)}; \phi_{\theta})$ with $q(\theta^{(\ell)}; \phi_{\theta})^\alpha$ in \eqref{eq:ELBO4} and reducing $\alpha$ from a large initial value to 1 during training.

\section{Forward Filter Recursion} \label{sec:forward_filter}
See \cite{West2006} for a general introduction to forward filtering algorithms for linear state-space models.
We adapt this as follows.
Upon applying the It\^o formula with the integrating factor $G(t,x) = x e^{\theta_1 t}$, the solution to \eqref{OU_SDE} can be obtained by
\begin{equation}
X_{t+\Delta t}  | \left(X_{t} = x_{t}\right) \sim N\left(x_{t} e^{\left(-\theta_1 \Delta t\right)} + \theta_2 \left(1-e^{-\theta_1 \Delta t}\right), \frac{\theta_3^2}{2 \theta_1} \left(1 - e^{-2\theta_1 \Delta t}\right)\right). \label{eq:OU_sol}
\end{equation}
Assuming $N$ observations on a regular grid of time-step $\Delta t = t_{i} - t_{i-1}$, the marginal parameter posterior is given by
\begin{equation}\label{eq:OU_target}
p(\theta|y_{t_0:t_N}) \propto p(\theta) p(y_{t_0:t_N}|\theta),
\end{equation}
where $p(y_{t_0:t_N}|\theta)$ is the marginal likelihood obtained from integrating out the latent variables from $p(\theta, x_{t_0:t_N}| y_{t_0:t_N})$. As can be seen from \eqref{eq:OU_sol}, the OU process is linear and Gaussian. Hence, for a Gaussian observation model \eqref{eq:obs_model}, the marginal likelihood is tractable and can be efficiently computed via a forward filter recursion. A forward filter recursion utilises the factorisation
\begin{equation}
p(y_{t_0:t_N} | \theta) = p(y_{t_0}| \theta) \prod_{i=1}^{N} p(y_{t_i} | y_{t_0:t_{i-1}}, \theta),
\end{equation}
by recursively evaluating each form. 

Assuming $x_{t_0} \sim N(a, c)$ a priori, we begin by calculating 
\begin{equation}
p(y_{t_0}|\theta) = N(y_{t_0}; a, c + \sigma^2).
\end{equation}
The posterior at $t_0$ is $x_{t_0} \big| y_{t_0}, \theta \sim N(a_0, c_0)$ with
\begin{align}
a_0 &= a + c(c+\sigma^2)^{-1}(y_{t_0}-a),\\
c_0 &= c - c(c+\sigma^2)^{-1} c.
\end{align}

Now suppose that $x_{t_i} | y_{t_0 : t:i} \sim N(a_i, c_i)$. The prior at time $t_{i+1}$ is therefore
\begin{equation}\label{eq:ff1}
x_{t_{i+1}} \big| y_{t_0:t_i} \sim N\left(a_i e^{-\theta_1 \Delta t } + \theta_2\left(1 - e^{-\theta_1 \Delta t}\right), \frac{\theta_3^2}{2\theta_1} \left(1 - e^{-2 \theta_1 \Delta t}\right) + c_i e^{-2\theta_1 \Delta t} + \sigma^2 \right),
\end{equation}
which, from the observation model \eqref{eq:obs_model}, gives us the one-step ahead forecast
\begin{equation}\label{eq:one_step}
y_{t_{i+1}} \big| y_{t_0:t_i}, \theta \sim N\left(a_i e^{-\theta_1 \Delta t } + \theta_2\left(1 - e^{-\theta_1 \Delta t}\right), \frac{\theta_3^2}{2\theta_1} \left(1 - e^{-2 \theta_1 \Delta t}\right) + c_i e^{-2\theta_1 \Delta t} + \sigma^2 \right).
\end{equation}
Hence the marginal likelihood can be recursively updated using
\begin{equation}\label{eq:ff2}
p(y_{t_0:t_{i+1}}| \theta) = p(y_{t_0:t_i} | \theta) p(y_{t_{1+i}} | y_{t_0:t_i}, \theta),
\end{equation}
where $p(y_{t_{i+1}} | y_{t_0:t_i}, \theta)$ is the corresponding density of \eqref{eq:one_step}.

The posterior at time $t_{i+1}$ is obtained as $x_{t_{i+1}} | y_{t_0:t_{i+1}} \sim N(a_{i+1}, c_{i+1})$ where
\begin{align}
a_{i+1} =& \, a_i e^{-\theta_1 \Delta t } + \theta_2\left(1 - e^{-\theta_1 \Delta t}\right) + \left( \frac{\theta_3^2}{2\theta_1} \left(1 - e^{-2 \theta_1 \Delta t}\right) + c_i e^{-2\theta_1 \Delta t} +
 \sigma^2\right) \nonumber \\
& \left(\frac{\theta_3^2}{2\theta_1} \left(1 - e^{-2 \theta_1 \Delta t}\right) + c_i e^{-2\theta_1 \Delta t} + \sigma^2 \right)^{-1} \left(y_{t+\Delta t} -  a_i e^{-\theta_1 \Delta t } - \theta_2\left(1 - e^{-\theta_1 \Delta t}\right)\right), \label{eq:ff3}\\
c_{i+1} =& \, \frac{\theta_3^2}{2\theta_1} \left(1 - e^{-2 \theta_1 \Delta t}\right) + c_i e^{-2\theta_1 \Delta t} - \left( \frac{\theta_3^2}{2\theta_1} \left(1 - e^{-2 \theta_1 \Delta t}\right) + c_i e^{-2\theta_1 \Delta t} \right) \nonumber \\
& \left(\frac{\theta_3^2}{2\theta_1} \left(1 - e^{-2 \theta_1 \Delta t}\right) + c_i e^{-2\theta_1 \Delta t} + \sigma^2 \right)^{-1} \left( \frac{\theta_3^2}{2\theta_1} \left(1 - e^{-2 \theta_1 \Delta t}\right) + c_i e^{-2\theta_1 \Delta t} \right).\label{eq:ff4}
\end{align}
Evaluation of \eqref{eq:ff1}-\eqref{eq:ff4} for $i = 0,1,\ldots,N-1$ gives the marginal likelihood $p(y_{t_0:t_N} | \theta)$. Finally, we note that the marginal parameter posterior $p(y_{t_0:t_N} | \theta)$ is intractable. Therefore, we sample \eqref{eq:OU_target} using a random walk Metropolis-Hastings scheme (see e.g. \cite{Golightly2015}).

\section{Implementation Details} \label{sec:imp_details}
For both examples we made use of the following hyperparameter settings:
\begin{itemize}
	\item $n=50$ Monte Carlo samples in our gradient estimate \eqref{eq:ELBO4}.
	\item $m = 5$ composed local IAFs.
	\item $k = 10$ receptive field size.
	\item Each neural network used 5 layers with 20 hidden units and rectified linear activation function.
	\item We used the Adam optimizer \cite{DBLP:journals/corr/KingmaB14} in Tensorflow to maximise \eqref{eq:ELBO3}.
\end{itemize}

We additionally took the final elementwise transformation $h$ to be the softplus function to ensure positivity in the SIR example. 

\end{document}